%% file: main.tex
\documentclass[journal,twoside,web]{ieeecolor}
\usepackage{generic}
\usepackage{cite}
\usepackage{amsmath,amssymb,amsfonts}
\usepackage{algorithmic}
\usepackage{graphicx}
\usepackage{algorithm,algorithmic}
\usepackage{hyperref}
\hypersetup{hidelinks=true}
\usepackage{textcomp}

\usepackage{amsmath,amsfonts}
\usepackage{algorithmic}
\usepackage{algorithm}
\usepackage{array}
\usepackage[caption=false,font=normalsize,labelfont=sf,textfont=sf]{subfig}
\usepackage{textcomp}
\usepackage{stfloats}
\usepackage{url}
\usepackage{verbatim}
\usepackage{graphicx}
\usepackage{cite}
\usepackage{xcolor}
\usepackage{multirow}
\usepackage{booktabs}
\usepackage{tabularx}
\usepackage{adjustbox}
\usepackage{hyperref}
\usepackage{float}
\usepackage{color}
\usepackage{soul}
\def\BibTeX{{\rm B\kern-.05em{\sc i\kern-.025em b}\kern-.08em
    T\kern-.1667em\lower.7ex\hbox{E}\kern-.125emX}}
\begin{document}
\title{CLAD-Net: Continual Activity Recognition in Multi-Sensor Wearable Systems}

\author{Reza Rahimi Azghan$^{\star}$,  Gautham Krishna Gudur$^{\dagger}$, Mohit Malu$^{\star}$ \\Edison Thomaz$^{\dagger}$, Giulia Pedrielli$^{\star}$, Pavan Turaga$^{\star}$, Hassan Ghasemzadeh$^{\star}$
\thanks{$^{\star}$Arizona State University, Phoenix, AZ, USA,\vfill$^{\dagger}$ The University of Texas at Austin, Austin, TX, USA}%
}

\maketitle

\input{Sections/00_abstract}
\input{Sections/01_introduction}

\input{Sections/02_related_work}
\input{Sections/03_proposed_method}
\input{Sections/04_results}
\input{Sections/05_conclusion}

\bibliographystyle{IEEEtran}
\bibliography{refs}

\end{document}

%% file: Sections/00_abstract.tex
\begin{abstract}
The rise of deep learning has significantly advanced human behavior monitoring using wearable sensor data. In particular, human activity recognition (HAR) through deep models has been extensively explored. However, most existing methods assume a stationary data distribution, an assumption that often breaks down in real-world settings. For instance, sensor data collected from one subject performing an activity may differ substantially in distribution from that of another subject. In continual learning settings, this shift is modeled as a sequence of tasks, where each task corresponds to a new subject. This makes the system susceptible to catastrophic forgetting, where previously learned patterns degrade as new ones are acquired. This limitation is particularly critical in healthcare monitoring systems, where reliable long-term patient activity tracking is essential for clinical decision support, rehabilitation monitoring, and elderly care applications. The forgetting problem is further exacerbated by the reliance on labeled data, which is often sparse or inconsistently available in studies involving human participants. To address these challenges, drawing inspiration from complementary learning systems, we introduce CLAD-Net—\underline{C}ontinual \underline{L}earning with \underline{A}ttention and \underline{D}istillation—an innovative framework that enables deep models in wearable sensor systems to be updated continuously without compromising performance on previously learned tasks. CLAD-Net combines a self-supervised transformer, serving as the system's long-term memory, with a supervised convolutional neural network (CNN) trained with knowledge distillation for activity classification. The transformer captures global activity patterns through cross-attention across body-mounted sensors and learns generalizable representations without requiring labels. In parallel, the CNN incorporates knowledge distillation to retain past knowledge during subject-wise fine-tuning. We evaluate CLAD-Net on three benchmark HAR datasets spanning 31 continual learning streams across varying numbers of subjects and demonstrate superiority over replay-free continual learning baselines, including Learning without Forgetting (LwF) and Elastic Weight Consolidation (EWC), while offering a competitive privacy-preserving alternative to replay-based methods such as Experience Replay (ER), ER-ACE, and DER++ that require storing user data. On PAMAP2, CLAD-Net achieves 80.78\% final accuracy with 11.05\% forgetting, while on DnSA it attains 81.14\% accuracy with 8.68\% forgetting, and on RealWorld it reaches 64.34\% accuracy with 24.34\% forgetting. Furthermore, in semi-supervised settings with only 10–20\% labeled data, CLAD-Net maintains superior performance, showcasing its robustness to label scarcity. Extensive ablation studies also confirm the contribution of each module to the system's overall effectiveness.
\end{abstract}




%% file: Sections/01_introduction.tex
\section{Introduction}

Adapting to dynamic environments is a fundamental characteristic of any intelligent system. Through years of evolution, humans have developed the ability to adjust to new situations without losing previously acquired knowledge. Ideally, AI systems should exhibit a similar capability: acquiring tasks sequentially while performing as if all tasks were learned simultaneously. Despite their remarkable capacity to learn from data streams, many systems still struggle with retaining past knowledge~\cite{wang2024comprehensivesurveycontinuallearning, FRENCH1999128}. Learning continually from data allows models to balance the stability-plasticity trade-off~\cite{Kirkpatrick_2017}, which in turn enables them to generalize to future tasks while preserving prior learning. However, in practice, models often prioritize plasticity over stability, leading to catastrophic forgetting, a phenomenon that becomes even more severe when new tasks differ significantly in distribution from previous ones~\cite{vandeven2019scenarioscontinuallearning}. This challenge is particularly acute in healthcare monitoring systems, where continuous exposure to distribution shifts is inherent to real-world deployment.

In healthcare applications, machine learning systems must continuously adapt to new patients while maintaining accurate recognition across all enrolled individuals. Unlike controlled laboratory settings, clinical data streams do not follow an i.i.d. process~\cite{Lu_2018, koh2021wildsbenchmarkinthewilddistribution}. Instead, each new patient introduces an inherent distribution shift, as individuals perform the same activities in distinct ways due to variations in age, mobility, health status, and movement patterns~\cite{Soleimani_2021, kumarsahHAR}. This is especially critical in applications such as post-stroke rehabilitation monitoring, fall risk assessment in elderly populations, and chronic disease management, where HAR systems must learn from diverse patient populations without forgetting previously learned patterns~\cite{pamap2_physical_activity_monitoring_231, sztyler2016realworld}. Traditional approaches that retrain from scratch on combined patient data are impractical due to privacy regulations, computational constraints, and the need for real-time deployment on resource-limited wearable devices. Moreover, storing and accessing historical patient data raises significant privacy concerns under healthcare regulations, making exemplar-free continual learning approaches essential for clinical deployment.

\begin{figure}
    \centering
    \includegraphics[width=\linewidth]{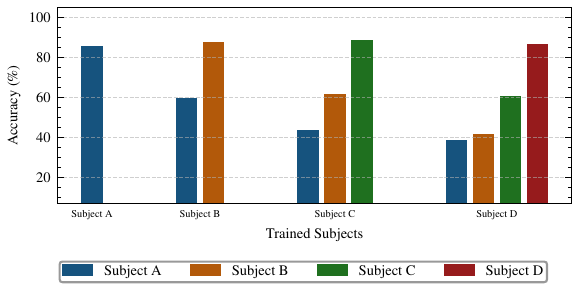}
    \caption{Fine-tuning the model on new subjects leads to a decline in accuracy on previously seen subjects}
    \label{fig:forgetting_intro}
\end{figure}

The severity of this challenge is illustrated in Fig.~\ref{fig:forgetting_intro}, where a 7-layer MLP is trained sequentially on the first four subjects of the PAMAP2 dataset~\cite{pamap2_physical_activity_monitoring_231}. As the model adapts to each new subject, its accuracy on previously seen subjects steadily declines, demonstrating the catastrophic forgetting phenomenon in a realistic HAR scenario. This pattern of degradation is typical when models encounter distribution shifts without mechanisms to preserve prior knowledge, resulting in either failure to generalize to new subjects or complete loss of learned decision boundaries for previous ones.

Additionally, most intelligent systems operate under the assumption that abundant labeled data is available for training. However, in real-world applications, this assumption is often unrealistic~\cite{nooshinssl, rezacudle, maryampgl}. Labeling is typically the responsibility of subjects themselves, who may not consistently follow the data collection protocol. In activity recognition tasks using sensor data, for instance, participants might forget to label the actions they perform, further complicating the downstream learning process.

To tackle these challenges, many research directions draw inspiration from how humans learn and retain knowledge~\cite{ hadsell2020embracing}. For example, the human mind often revisits past experiences to reinforce memory, an idea reflected in Experience Replay (ER)~\cite{chaudhry2019tinyepisodicmemoriescontinual, lin1992self, liu2020mnemonics}. Additionally, humans naturally retain prior knowledge while learning new skills. The brain preserves learned representations even as it acquires new information, much like distillation-based methods~\cite{zhang2025continualdistillationlearningknowledge, usmanova2021distillationbasedapproachintegratingcontinual, li2017learningforgetting}, which helps neural networks retain responses to earlier tasks during adaptation to new ones. Furthermore, the brain employs distinct mechanisms for long-term and short-term memory, using complementary systems~\cite{mcclelland1995there, pham2021dualnetcontinuallearningfast} to encode new experiences while preserving past knowledge—a principle echoed in models like DualNet~\cite{pham2021dualnetcontinuallearningfast} and Kaizen~\cite{tang2024kaizenpracticalselfsupervisedcontinual}.

Humans also excel at learning from unstructured, unlabeled data. Infants, for instance, recognize objects and speech patterns by predicting and associating outcomes from raw sensory input~\cite{Clark_2013, grill2020bootstrap, hashemi2025ultra}. This ability to derive learning signals from context aligns with self-supervised learning, where models extract useful representations through pretext tasks without labeled data~\cite{rezacudle}. Although many self-supervised methods have been developed to address label scarcity, their role in improving knowledge retention for downstream classifiers remains an open question~\cite{tang2024balancingcontinuallearningfinetuning}.

This work makes three key contributions to address these challenges. First, we tackle domain-incremental learning in the context of HAR, where the model continuously encounters time-series data from new subjects and must generalize to their activity patterns while retaining knowledge of previously seen subjects. Second, we introduce CLAD-Net, a two-component architecture specifically designed to reduce catastrophic forgetting. At its core is a self-supervised transformer that employs cross-attention across sensor readings from different body parts, learning generalizable representations without requiring labels. This is coupled with a supervised CNN trained using knowledge distillation to preserve past knowledge during subject-wise fine-tuning. This combination addresses both the challenges of catastrophic forgetting and limited label availability, a scenario increasingly common in real-world deployments. Third, we conduct extensive experiments on three benchmark activity recognition datasets, demonstrating that CLAD-Net outperforms replay-free continual learning baselines (LwF and EWC) while maintaining competitive performance with replay-based methods without storing any exemplars. In label-scarce scenarios, CLAD-Net demonstrates superior forgetting resistance compared to all baseline methods.







%% file: Sections/02_related_work.tex
\section{Related Work}

HAR using wearable sensors has become increasingly important in healthcare applications, enabling continuous patient monitoring in both clinical and home environments~\cite{bolatov2024glula, convboost}. The task involves detecting and classifying physical activities such as walking, standing, running, or cycling based on time-series data collected from subjects and their environment, often through inertial sensors like accelerometers and gyroscopes. These systems support various medical applications, including fall detection for elderly care, rehabilitation progress tracking for stroke patients, and activity-based intervention for chronic disease management. In recent years, deep learning approaches, particularly convolutional neural networks (CNNs), have shown strong performance in extracting discriminative features from sensor data~\cite{bolatov2024glula, kumarsahHAR, abdullahsensor, convboost}, achieving remarkable results in controlled laboratory settings.

However, most existing HAR systems face significant challenges when deployed in real-world healthcare environments. First, they require extensive labeled data from each new patient, yet clinical settings often provide limited or inconsistently available labels due to the burden on medical staff and patients~\cite{kumarsahHAR, SCHIEMER2023101817}. Second, these models cannot adapt to individual movement patterns without complete retraining, making them impractical for resource-limited wearable devices. Third, and most critically, they are prone to catastrophic forgetting when exposed to new subjects or data distributions, lacking mechanisms to preserve performance across diverse patient populations~\cite{Jha_2021, SCHIEMER2023101817, kumarsahHAR}. Moreover, privacy concerns in medical settings necessitate methods that avoid storing raw patient sensor data, making replay-based continual learning approaches problematic for clinical deployment.

To address label scarcity, recent self-supervised approaches~\cite{yuan2024sslhar, nooshinssl, rezacudle} have demonstrated the value of large-scale pretraining on unlabeled sensor data to improve robustness and generalization, reducing the annotation burden in healthcare applications. However, the role of self-supervised learning in mitigating catastrophic forgetting while maintaining performance across sequential patient enrollments remains underexplored. Despite these advances, effectively addressing both forgetting and label scarcity simultaneously in HAR for healthcare monitoring remains an open and important problem~\cite{pham2021dualnetcontinuallearningfinetuning, tang2024balancingcontinuallearningfinetuning}.

Continual learning, which aims to overcome the problem of catastrophic forgetting, has traditionally been studied somewhat separately from HAR. It focuses on training models under non-stationary data distributions, where new tasks are presented sequentially over time~\cite{Kirkpatrick_2017, rebuffi2017icarlincrementalclassifierrepresentation}. This paradigm is often referred to as lifelong learning or incremental learning in the literature, with the terms being used interchangeably in most cases~\cite{wang2024comprehensivesurveycontinuallearning, PARISI201954}. The core goal is to enable models to adapt to new tasks or data domains without forgetting previously learned knowledge—a capability essential for real-world HAR systems that operate over extended periods or across diverse users. Continual learning research is typically categorized into three main scenarios~\cite{van2022three, vandeven2019scenarioscontinuallearning}:
\begin{itemize}
    \item Domain-incremental learning: Tasks share the same label space but differ in data distribution—for example, recognizing the same activities across different users or sensor setups.
    \item Task-incremental learning: Each task has a distinct label space and distribution, and task identities are provided during both training and evaluation.
    
    \item Class-incremental learning: Like task-incremental learning, but task identities are only available during training. 
    
\end{itemize}

Many continual learning approaches are inspired by how the human brain handles forgetting. Methods such as ~\cite{arani2022learningfastlearningslow} and ~\cite{gomezvilla2023plasticityoptimizedcomplementarynetworksunsupervised} follow the idea of complementary learning systems, separating short-term and long-term memory components within the architecture. In contrast, simpler strategies like Experience Replay (ER) use memory buffers that store samples from previous tasks to help the model retain past knowledge during training on new data~\cite{chaudhry2019tinyepisodicmemoriescontinual}. Other approaches, like ~\cite{szatkowski2023adaptteacherimprovingknowledge} and ~\cite{li2017learningforgetting}, adopt knowledge distillation, penalizing the model when its parameters deviate significantly while learning new tasks. While much of the existing research has focused on image data, interest in continual learning for time-series data—especially in HAR—has grown in recent years. ~\cite{Jha_2021} evaluated a range of continual learning approaches, including both regularization-based techniques and rehearsal methods, on several HAR time-series datasets. In another study, ~\cite{YIN2023171} proposed an attentive recurrent neural network to combat catastrophic forgetting in HAR, showing that incorporating attention mechanisms into lightweight models can effectively support continual learning.
\vspace*{-.12em}

Transformers have recently emerged as powerful models for sensor-based HAR, largely inspired by the success of Vision Transformers (ViTs) in computer vision, which model spatial relationships by treating image patches as sequences~\cite{dosovitskiy2021an, elhambakhsh2025domainadaptationlargelanguage, chen2021crossvitcrossattentionmultiscalevision}. In wearable HAR, a similar principle applies—sensor readings over time or across body locations can be treated as sequences or segments for attention-based modeling. A core innovation in this domain is the use of cross-attention mechanisms to capture dependencies between distinct body parts, sensor modalities, or temporal patterns. For instance, ~\cite{pang2024cross} proposed a cross-attention enhanced pyramid network that establishes interrelationships across sensor channels, time steps, and feature dimensions, enabling the model to suppress noise and amplify activity-relevant signals. Similarly, ~\cite{tang2022triple} introduced a triple-attention framework that performs attention across sensor, temporal, and channel axes, showing that cross-dimensional fusion yields significant improvements in HAR performance. ~\cite{pramanik2023reverse} took a top-down approach by proposing a "reverse attention" mechanism, where global contextual features attend back to individual sensor streams. ~\cite{xiao2022twostream} presented a two-stream transformer architecture that separately models temporal dynamics and cross-sensor spatial relationships, fusing them through attention-weighted integration to account for both intra-sensor sequences and inter-sensor configurations. Collectively, these models demonstrate the growing role of cross-attention in HAR and offer a principled way to model complex spatial-temporal patterns across the human body using wearable sensor data.

%% file: Sections/03_proposed_method.tex
\section{Proposed Approach}

\begin{figure*}
    \centering
    \includegraphics[width=\linewidth]{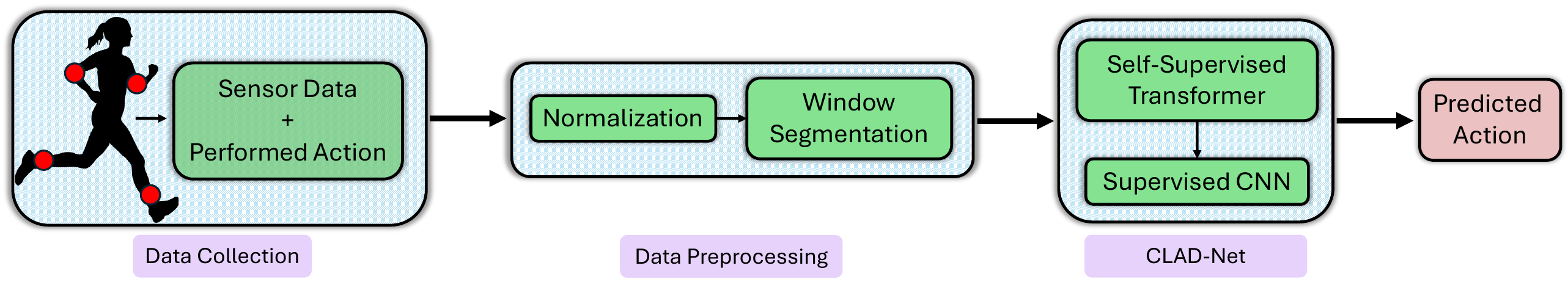}
    \caption{The complete pipeline of the proposed system, composed of three core components: data collection, data preprocessing, and the final model, CLAD-Net.}
    \label{fig:system}
\end{figure*}

\subsection{System overview}

In this section, we describe the components of our proposed system, following the overall flow from data sensing to model transfer and evaluation. As illustrated in Fig. \ref{fig:system}, the system is composed of three main components.

The first component is the data collection module, where inertial sensors are installed on various parts of the subject’s body as they perform a predefined set of activities. Subjects are also asked to provide the system with activity labels to indicate the type of activity they were performing at any given instant.  

The second component is the data fetching and preprocessing module. Here, the raw sensor data collected from each subject is aggregated and normalized on a per-subject basis. A window segmentation algorithm is then applied to divide the continuous sensor streams into fixed-length segments. Each window is assigned a label if the corresponding activity is provided by the subject. Formally, each labeled window can be represented as $(\mathbf{x},y)$, where $\mathbf{x}\in \mathbb{R}^{l\times d} $, with $l$ being the window length and $d$ the number of sensor channels. Since our experiments include a semi-supervised learning setup, labels $y$ may not be available for every window. 

The third and final component is the proposed machine learning model, which is designed to classify input windows into their corresponding activity labels. The main goal of the model is to maintain high classification accuracy for each new subject while minimizing the loss of information about previously seen subjects. To this end, we introduce a two-part architecture optimized with distinct objective functions, which are described in the following subsections.

\subsection{Theoretical Background}
In this work, we formulate the problem of continual learning in the application of human activity recognition as a domain-incremental learning paradigm. At each task $t\in \{1, ..., T\}$, the system receives a batch of input-label pairs $\mathcal{D}_{t}=\{(\mathbf{x}_{i, t}, y_{i, t})\}_{i=1}^{N_t}$, where $\mathbf{x}_{i, t}\in \mathbb{R}^{l\times d}$ denotes a multivariate time-series of length $l$ with $d$ sensor channels (e.g., accelerometer, gyroscope), and $y_{i, t}\in \mathcal{Y}$ is the corresponding activity label. $T$ is the number of overall subjects, and each task $t$ corresponds to a distinct subject associated with a domain-specific distribution $\mathcal{D}_{t}\sim \mathbf{P}_{t}(\mathbf{x}, y)$.

Under the domain-incremental learning setting, we assume a static label distribution across tasks, i.e.,

$$\mathbf{P}_{t}(y)=\mathbf{P}_{t+1}(y), \hspace{1em} \forall t\in \{1, ..., T-1\}$$
but allow the marginal distribution over the time-series data to vary, such that

$$\mathbf{P}_{t}(\mathbf{x})\neq \mathbf{P}_{t'}(\mathbf{x}), \hspace{1em} \forall t\neq t'$$

Like in real settings, CLAD-Net is exposed to a continuum of the data points $\{(\mathbf{x}_{i, t}, y_{i, t}, t)\}$, where the distribution can be changed from $\mathbf{P}_{t}$ to $\mathbf{P}_{t+1}$ at any instant. Also, to reflect the constraints of real-world, memory-bound systems, the model does not have access to data from previous domains.

\subsection{CLAD-Net}

CLAD-Net is composed of two key modules, as illustrated in Fig. \ref{fig:system}. The first is a self-supervised transformer that functions as the architecture’s long-term memory. This component operates independently of the task (subject) identifier and activity label, and only relies on dynamically distributed input variables. 

The second module is a CNN-based classifier that processes the raw time-series data along with their corresponding labels. During training and inference, the representation learned by the transformer is concatenated with the classifier’s final feature layer. This knowledge consolidation allows the classifier to benefit from the transformer’s long-term representations. 

Both the self-supervised transformer and  the supervised CNN are trained sequentially as each new subject arrives. Specifically, when training on subject $t$, the transformer receives only data from subject $t$ and applies the objective function on augmented views of that subject's samples. No data from future subjects ($t'>t$) is used during training, ensuring strict adherence to the continual learning protocol without data leakage. The transformer learns representations online as new subjects stream in, with no access to future subject distributions.

Further details on the architecture of both modules, their forward passes, and training procedures are presented in the following subsections.

\begin{figure}
    \centering
    \includegraphics[width=\linewidth]{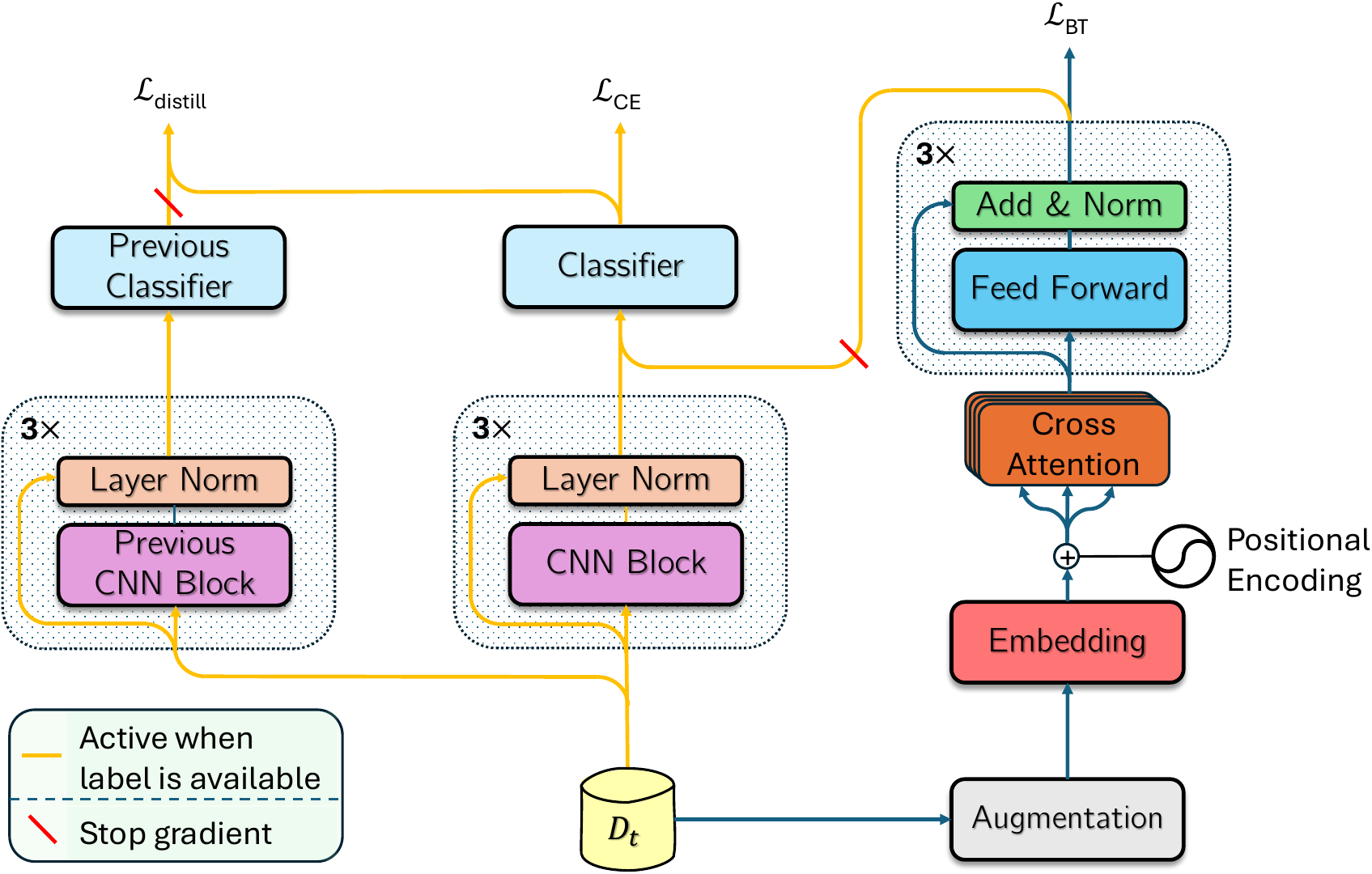}
    \caption{Overview of CLAD-Net, including its architectural components and training workflow.}
    \label{fig:model}
\end{figure}

\subsubsection{Self-supervised Transformer}
Fig. \ref{fig:model} illustrates the backbone architecture of the self-supervised transformer, which serves as the long-term memory of CLAD-Net. The standard transformer architecture is typically made up of an encoder-decoder for data reconstruction. We adapt this design by retaining only the encoder and framing it as a self-supervised learning task .

Let $\mathbf{x} \in \mathbb{R}^{l \times d}$ represent the input time-series segment, where $d$ is the total number of sensor channels and $l$ is the sequence (window) length. In many sensor-based HAR systems, sensors are positioned on multiple distinct body regions. To exploit this structure, we assume that the $d$ sensor channels are grouped into $n$ non-overlapping subsets, each associated with a specific body part. We denote this partition as ${\mathbf{x}^{(1)}, \mathbf{x}^{(2)}, \dots, \mathbf{x}^{(n)}}$, where $\mathbf{x}^{(i)} \in \mathbb{R}^{l \times d_i}$ and $\sum_{i=1}^{n} d_i = d$. This grouping allows the model to encode localized representations for each body region, rather than treating the input as a flat sequence of sensor channels. These body-part-specific segments are then processed independently in downstream modules.

Subsequently, each body-part segment is passed through an encoder to extract localized features. This step is analogous to Vision Transformers~\cite{dosovitskiy2021an}, where image patches are projected into a shared embedding space. Formally, for each body part, the encoder is a parameterized learnable function $f_{\text{embed}, i}:\mathbb{R}^{l\times d_i}\rightarrow\mathbb{R}^{l\times d_\text{model}}$ mapping raw sensor inputs to a fixed-dimensional latent space. The resulting embeddings are then enriched with temporal information using a fixed sinusoidal positional encoder, $g_{\text{pos}}: \mathbb{R}^{l\times d_{\text{model}}}\rightarrow\mathbb{R}^{l\times d_\text{model}}$~\cite{vaswani2017attention}. This step allows the model to distinguish events not just by what occurred, but also by when they occurred. The encoded and temporally-aware segments are then ready to be processed by the attention mechanism that follows.

At this stage, we have $n$ temporally encoded body-part representations denoted as $\{\mathbf{z}^{(1)}, \dots, \mathbf{z}^{(n)}\}$, where each $\mathbf{z}^{(i)} = g_{\text{pos}}(f_{\text{embed}}(\mathbf{x}^{(i)}))$ represents the embedded and positionally encoded signal from the $i$-th body region.

To capture interdependencies between body parts, we apply an $n$-branch multi-head cross-attention mechanism across these representations. Each branch computes the attention between a shared query signal $\mathbf{z}^{(q)}$ (from a designated reference body part) and a distinct body-part representation $\mathbf{z}^{(i)}$ used as both key and value. This setup allows the query region to selectively incorporate contextual information from other body regions.

The attention in each branch is computed using the standard scaled dot-product attention:
\begin{equation}
\text{Attention}(\mathbf{Q}, \mathbf{K}, \mathbf{V}) = \text{Softmax} \left( \frac{\mathbf{Q} \mathbf{K}^\top}{\sqrt{d_{\text{model}}}} \right) \mathbf{V}.
\end{equation}

For the $i$-th cross-attention branch with $m_H$ heads, we compute:
\begin{equation}
\text{CrossAttn}(\mathbf{z}^{(q)}, \mathbf{z}^{(i)}) = \left[ \mathbf{H}^{(i)}_1, \dots, \mathbf{H}^{(i)}_{m_H} \right] \mathbf{W}^{(i)}_\mathbf{H},
\end{equation}

Where each head output is defined as:
\begin{equation}
\mathbf{H}^{(i)}_h = \text{Attention} \left( \mathbf{z}^{(q)} \mathbf{W}^{(i)}_{\mathbf{Q}_h},\ \mathbf{z}^{(i)} \mathbf{W}^{(i)}_{\mathbf{K}_h},\ \mathbf{z}^{(i)} \mathbf{W}^{(i)}_{\mathbf{V}_h} \right).
\end{equation}

Here:
\vspace{1em}
\begin{itemize}
  \item $\mathbf{W}^{(i)}_{\mathbf{Q}_h},\ \mathbf{W}^{(i)}_{\mathbf{K}_h},\ \mathbf{W}^{(i)}_{\mathbf{V}_h} \in \mathbb{R}^{d_{\text{model}} \times d_{\text{model}}}$ are the learnable projection matrices for the $h$-th head in the $i$-th branch.
  \vspace{1em}
  \item $\mathbf{W}^{(i)}_\mathbf{H} \in \mathbb{R}^{(m_H \cdot d_{\text{model}}) \times d_{\text{model}}}$ is the output projection matrix that maps the concatenated head outputs back to the original model dimension.
\end{itemize}

This multi-branch structure enables the model to retain a fixed query representation while attending to multiple body-part signals in parallel. The outputs from all attention branches are then aggregated and forwarded to the subsequent layers of the transformer architecture.

The final stage of the transformer comprises three stacked feedforward blocks, each followed by a residual connection and layer normalization. The process begins by normalizing the cross-attention output before passing it to the first feedforward layer. The output of this layer is added back to its input, followed by another layer normalization and a second feedforward transformation. This pattern is repeated once more, yielding a total of three feedforward layers interleaved with Add \& Norm operations. 

As discussed earlier, the transformer is trained in a self-supervised manner using a contrastive objective. Although several loss functions can be employed in this framework, we adopt the BYOL (Bootstrap Your Own Latent) loss~\cite{grill2020bootstrap} following its successful application in continual learning scenarios. To compute this loss, we generate two distinct augmentations of the input sample $\mathbf{x}$, denoted as $\mathbf{\bar{x}}$ and $\mathbf{\hat{x}}$. These augmented views are created using time-series-specific transformations, such as additive Gaussian noise or zero masking. We evaluate the effect of different augmentation strategies empirically in the results section. Each augmented input is passed independently through the transformer $\Psi$, producing the output representations $\mathbf{\bar{r}} = \Psi(\mathbf{\bar{x}})$ and $\mathbf{\hat{r}} = \Psi(\mathbf{\hat{x}})$. The BYOL loss is then computed by maximizing the cosine similarity between the $\ell_2$-normalized representations:
\begin{equation}
\mathcal{L}_{\text{BYOL}} = -\frac{1}{B}\sum_{b=1}^{B} \frac{\bar{\mathbf{r}}_b \cdot \hat{\mathbf{r}}_b}{\|\bar{\mathbf{r}}_b\|_2 \cdot \|\hat{\mathbf{r}}_b\|_2},
\end{equation}
where $B$ is the batch size, and $\bar{\mathbf{r}}_b$ and $\hat{\mathbf{r}}_b$ refer to the representations of the $b$-th sample from the two augmented views. The loss function encourages the two views to produce similar representations while avoiding collapse through the asymmetric architecture and momentum-based updates inherent to BYOL. Algorithm \ref{alg:transformer_ssl} summarizes the forward and training procedure of the transformer model.

\begin{algorithm}
\caption{Self-Supervised Transformer Training with BYOL Loss}
\label{alg:transformer_ssl}
\begin{algorithmic}[1]
\REQUIRE Input time-series batch $\{\mathbf{x}_b\}_{b=1}^B$, augmentations $\mathcal{A}_1$, $\mathcal{A}_2$, body partitioning function $\mathcal{P}$, optimizer $\mathcal{O}$
\vspace{0.5em}
\STATE \textbf{Augment input samples:}
\FOR{$b = 1$ to $B$}
    \STATE $\bar{\mathbf{x}}_b \gets \mathcal{A}_1(\mathbf{x}_b)$
    \STATE $\hat{\mathbf{x}}_b \gets \mathcal{A}_2(\mathbf{x}_b)$
\ENDFOR
\vspace{0.5em}
\STATE \textbf{Forward pass through transformer:}
\FOR{$b = 1$ to $B$}
    \FOR{$s \in \{\bar{\mathbf{x}}_b,\ \hat{\mathbf{x}}_b\}$}
        \STATE Partition input into $n$ body parts: $\{ \mathbf{x}^{(i)} \}_{i=1}^n \gets \mathcal{P}(s)$\vspace{-1.2em}
        \FOR{$i = 1$ to $n$}
            \STATE $\mathbf{z}^{(i)} \gets g_{\text{pos}}(f_{\text{embed}, i}(\mathbf{x}^{(i)}))$
        \ENDFOR
        \STATE Select query part $\mathbf{z}^{(q)}$ (e.g., hand)
        \FOR{$i = 1$ to $n$}
            \STATE $\mathbf{a}^{(i)} \gets \text{CrossAttn}(\mathbf{z}^{(q)}, \mathbf{z}^{(i)})$
        \ENDFOR
        \STATE $\mathbf{a} \gets \text{Aggregate}(\{\mathbf{a}^{(i)}\}_{i=1}^n)$
            \STATE $\mathbf{a} \gets \text{FF}_1(\text{Norm}_1(\text{Dropout}(\mathbf{a})))$
            \STATE $\mathbf{a} \gets \text{FF}_2(\text{Norm}_2(\mathbf{a} + \text{Dropout}(\mathbf{a})))$
            \STATE $\mathbf{r}_b \gets \text{FF}_3(\text{Norm}_3(\mathbf{a} + \text{Dropout}(\mathbf{a})))$
    \ENDFOR
\ENDFOR
\vspace{0.5em}
\STATE \textbf{Compute BYOL loss:}
\STATE Normalize outputs: $\bar{\mathbf{r}}, \hat{\mathbf{r}} \in \mathbb{R}^{B \times d_{\text{model}}}$
\STATE Compute cosine similarity and loss $\mathcal{L}_{\text{BYOL}}$
\vspace{0.5em}
\STATE \textbf{Backward pass:}
\STATE Compute gradients: $\nabla_\theta \mathcal{L}_{\text{BYOL}}$
\STATE Update model parameters: $\theta \gets \mathcal{O}(\theta, \nabla_\theta \mathcal{L}_{\text{BYOL}})$
\end{algorithmic}
\end{algorithm}
\subsubsection{Supervised Convolutional Neural Network}

In parallel with the self-supervised transformer, our system includes a supervised CNN designed for the downstream task of activity classification. Unlike the transformer, which operates independently of labels and task identity, the CNN is trained in a fully supervised and task-aware setting—meaning it has access to both activity labels and the subject identifier during training. This allows the CNN to optimize directly for subject-specific decision boundaries.

The CNN architecture comprises three convolutional blocks. Each block contains four layers of 1D convolutions, followed by an average pooling layer and a nonlinearity. We add a residual connection and apply layer normalization after every block, resembling the Add \& Norm structure used in transformer and ResNet architectures. The final output of the CNN is concatenated with the transformer’s final representation vector, and the combined feature is passed to a linear classifier for prediction.

We train the CNN using a knowledge distillation\cite{farahmand2025hybridattentionmodelusing} approach inspired by Learning without Forgetting (LwF)~\cite{li2017learningforgetting}. Specifically, after training on the final batch of subject $t-1$, we store a frozen copy of the model’s parameters, denoted as $\theta_{t-1}$. When training on the subsequent subject $t$, we minimize a combined loss function that balances prediction accuracy and output consistency with the previous model: 
\begin{equation} 
\mathcal{L}_{\text{total}} = \mathcal{L}_{\text{CE}}(f_{\theta_{t}}(\mathbf{x}),\ y) + \lambda_{\text{distill}} \cdot \mathcal{L}_{\text{distill}}(f_{\theta_{t}}(\mathbf{x}),\ f_{\theta_{t-1}}(\mathbf{x}))
\end{equation} 

Where $\mathbf{x}$ and $y$ are the input-label pairs from subject $t$, and $\lambda_{\text{distill}}$ is a hyperparameter that controls the trade-off between classification accuracy and knowledge preservation. The distillation term is defined as the Kullback–Leibler (KL) divergence between the output distributions of the current model and those of the frozen model. Note that we do not store or replay any data from previous subjects. The only retained information is the snapshot of the model parameters $\theta_{t-1}$ after training on subject $t-1$. Algorithm \ref{alg:cnn_transformer} summarizes the forward process and the training procedure of the CNN module.

\begin{algorithm}
\caption{Forward and Training Procedure of Supervised CNN with Transformer Fusion}
\label{alg:cnn_transformer}
\begin{algorithmic}[1]
\REQUIRE Input time-series segment $\mathbf{x} \in \mathbb{R}^{l \times d}$, transformer output $\mathbf{r} \in \mathbb{R}^{l \times d_{\text{model}}}$, ground truth label $y$, model parameters $\theta$, previous model parameters $\theta_{t-1}$
, loss weight $\lambda$, KL temperature $\tau$, optimizer $\mathcal{O}$

\vspace{0.5em}
\STATE Initialize feature tensor: $\mathbf{h} \gets \mathbf{x}$

\vspace{0.5em}
\FOR{$i = 1$ to $3$} 
    \STATE $\mathbf{h}_{\text{block}} \gets \text{CNNBlock}_i(\mathbf{h})$
        \STATE $\mathbf{h} \gets \text{LayerNorm}(\mathbf{h} + \mathbf{h}_{\text{block}})$ 
\ENDFOR

\vspace{0.5em}
\STATE \textbf{Fuse with transformer output:}
\STATE $\mathbf{o} \gets \text{Concat}(\mathbf{h},\ \mathbf{r})$

\vspace{0.5em}
\STATE \textbf{Classification:}
\STATE $\hat{y} \gets \text{LinearClassifier}(\mathbf{o})$

\vspace{0.5em}
\STATE \textbf{Compute Total Loss:}
\STATE $\mathcal{L}_{\text{CE}} \gets \text{CrossEntropy}(\hat{y},\ y)$
\IF{previous model $f_{\theta_{t-1}}$ exists}
    \STATE $\mathcal{L}_{\text{distill}} \gets \text{KL}\left(\text{Softmax}\left(\frac{f_{\theta}(x)}{\tau}\right) \parallel \text{Softmax}\left(\frac{f_{\theta_{t-1}}(x)}{\tau}\right)\right)$

    \STATE $\mathcal{L}_{\text{total}} \gets \mathcal{L}_{\text{CE}} + \lambda \cdot \mathcal{L}_{\text{distill}}$
\ELSE
    \STATE $\mathcal{L}_{\text{total}} \gets \mathcal{L}_{\text{CE}}$
\ENDIF

\vspace{0.5em}
\STATE \textbf{Backward pass:}
\STATE Compute gradients $\nabla_\theta \mathcal{L}_{\text{total}}$
\STATE Update model parameters: $\theta \gets \mathcal{O}(\theta, \nabla_\theta \mathcal{L}_{\text{total}})$

\end{algorithmic}
\end{algorithm}

\subsubsection{Theoretical Motivation}

The combination of self-supervised learning and knowledge distillation addresses catastrophic forgetting through complementary stability mechanisms that target different aspects of the learning system.

\textbf{SSL for representation stability:} The self-supervised transformer learns representations by optimizing for invariance to semantic-preserving augmentations through predictive learning (BYOL objective) rather than subject-specific discriminative features. Formally, let $\mathcal{T}$ be the space of augmentations. The SSL objective:

\begin{equation} \min_{\theta_{\text{SSL}}} \mathbb{E}_{x \sim \mathcal{D}_t, \tau_1,\tau_2 \sim \mathcal{T}} \left[ \mathcal{L}_{\text{BYOL}}(f_{\theta_{\text{SSL}}}(\tau_1(x)), f_{\theta_{\text{SSL}}}(\tau_2(x))) \right] \end{equation}

encourages learning activity-level features invariant to subject-specific variations (motion speed, sensor placement), which naturally provides cross-subject generalization. This stands in contrast to supervised learning, which can overfit to subject-specific patterns and suffer catastrophic forgetting when subjects change.

\textbf{KD for decision boundary stability:} Knowledge distillation constrains the classifier's output distribution to remain consistent with previous subjects via: 

\begin{equation} \min_{\theta} \mathbb{E}_{x \sim \mathcal{D}_t} \left[ \mathcal{L}_{\text{CE}}(f_{\theta}(x), y) + \lambda \text{KL}(f_{\theta}(x) \parallel f_{\theta_{t-1}}(x))\right] \end{equation} 

This regularizes decision boundaries to avoid sudden shifts that would degrade performance on previous subjects, even without replaying their data.

Combining both mechanisms provides orthogonal stability: SSL stabilizes the feature space across subjects, while KD stabilizes the classifier's decision boundaries. This dual-level stability is more effective than either mechanism alone, as demonstrated in our ablation studies

%% file: Sections/04_results.tex
\section{Experiments}
In this section, we begin by introducing the datasets used to evaluate our proposed system, followed by a description of the employed evaluation metrics. We then outline the implementation details of our framework and present a series of experiments, including ablation studies, to assess the effectiveness of each component.
\subsection{Benchmarks}
\subsubsection{PAMAP2 Physical Activity Monitoring~{\normalfont\cite{pamap2_physical_activity_monitoring_231}}} 
The dataset consists of inertial sensor recordings from 8 subjects, each performing 12 predefined activities. An additional set of 6 optional activities was included in the original dataset, but these were excluded from our analysis due to inconsistent subject participation. IMU sensors were positioned on three body regions: the dominant hand, chest, and dominant ankle. Each region was equipped with three sensor modalities: accelerometer, gyroscope, and magnetometer, resulting in a total of 9 sensor channels per location. All sensors were sampled at 100 Hz. 

\subsubsection{Daily and Sports Activities~{\normalfont\cite{daily_and_sports_activities_256}} (DnSA)} 
This dataset includes sensor recordings from 8 subjects performing 18 distinct activities. IMU sensors were installed on five body regions: the torso, left arm, right arm, left leg, and right leg. For consistency with PAMAP2, we use data from three locations: the torso (chest), right arm (hand), and right leg (ankle). Each body part was equipped with an accelerometer, gyroscope, and magnetometer, resulting in a total of 9 sensor channels per location. All sensors were sampled at a frequency of 25 Hz.

\subsubsection{RealWorld HAR~{\normalfont\cite{sztyler2016realworld}}}
This dataset contains recordings from 15 subjects performing 8 activities: walking, jogging, sitting, standing, lying, climbing stairs up, climbing stairs down, and jumping. Each subject wore sensors on seven body locations simultaneously, sampled at 50 Hz. To maintain consistency with PAMAP2 and DnSA, we utilize data from three body locations: forearm (as a proxy for hand), chest, and shin (as a proxy for ankle). Forearm and shin placements capture functionally equivalent motion patterns to hand and ankle for activity recognition. Each location provides accelerometer, gyroscope, and magnetometer readings, yielding 9 sensor channels per location.

\subsection{Evaluation Metrics}
\subsubsection{Final Accuracy}
This metric quantifies the model's ability to maintain performance across all subjects after sequential training is complete and reflects how well knowledge is preserved in the final model state.
Mathematically:
\begin{equation}
    \text{FA} = \dfrac{1}{T}\sum_{t\in \{1,\cdots, T\}}A_{t, T}
\end{equation}
where $A_{t, T}$ is the accuracy on subject $t$ after training on all $T$ subjects.
\subsubsection{Forgetting Measure} 
This metric captures the severity of catastrophic forgetting. It quantifies the stability of previously acquired knowledge by measuring the decline in performance on earlier subjects. Mathematically:
\begin{equation}
    \text{FM} = \dfrac{1}{T}\sum_{t\in \{1,\cdots,T\}}\max _{t'\in\{1,\cdots,T\}} A_{t, t'} - A_{t, T}
\end{equation}

where $A_{t,t'}$ is the accuracy on subject $t$ after training on the subject $t'$.

\subsubsection{Learning Accuracy} This metric evaluates the model's plasticity by measuring how effectively it acquires new knowledge when first presented with each subject.
Mathematically:
\begin{equation}
\text{LA} = \dfrac{1}{T}\sum _{t\in \{1,\cdots,T\}}A_{t, t}    
\end{equation}
where $A_{t, t}$ is the accuracy on subject $t$ immediately after training on subject $t$.
\subsection{Implementation Details}

For all three datasets, we applied an overlapping window segmentation approach using a window size of 2 seconds and a 50\% overlap between successive windows. From each subject's segmented data, 80\% was used for training and the remaining 20\% for testing. To account for inter-subject variability in sensor measurements, we performed per-subject standardization by normalizing the data to zero mean and unit variance. Across all benchmarks, we use data from three body locations with three sensor modalities (accelerometer, gyroscope, and magnetometer) per location, resulting in 27 input channels total. In all experiments, the dominant hand (or forearm for RealWorld) was used as the query input for the cross-attention mechanism, while the key and value inputs included all body parts, including the hand.

The transformer was trained using a zero-masking strategy, where 10-30\% of the input time-series segment was randomly masked with zeros across 1-3 contiguous blocks. The embedding module consisted of a linear projection, followed by a sinusoidal positional encoding to inject temporal information. Each feedforward block comprised two linear layers with ReLU activation. The model was optimized using the BYOL loss over 50 epochs with a learning rate of 0.001.

For the supervised CNN module, we used three stacked convolutional blocks, each containing four convolutional layers followed by average pooling and a ReLU activation. The final classification layer was a linear projection. The CNN was trained for 50 epochs using the Adam optimizer with a learning rate of 0.001 and incorporating a distillation loss (with $\lambda_{\text{distill}}=1$) when parameters from the previous frozen model were available. 

All hyperparameters were selected through a grid search on a held-out validation split (10\% of training data) from all subjects. This validation data was not used during continual learning experiments. For the transformer architecture, we set the embedding dimension $d_{\text{model}}=256$, the number of attention heads $m_H=8$, and the feedforward dimension to 300. We used a dropout rate of 0.1 across all layers. For the knowledge distillation component, we set the distillation weight $\lambda_{\text{distill}}=1.0$ and temperature $\tau=2.0$. The CNN employed three convolutional blocks with 300 hidden channels and a dropout rate of 0.1. Both models used a batch size of 32 and were trained for 50 epochs per subject with the Adam optimizer. Learning rates were set to 0.001 for both the transformer and CNN. We experimented with learning rates in $\{0.0001, 0.0005, 0.001, 0.005\}$, distillation weights in $\{0.5, 1.0, 2.0\}$, and temperatures in $\{1, 2, 4\}$. The selected configuration provided the best balance between final accuracy and forgetting measure on the validation set.

\subsection{Results on benchmarks}
We compared CLAD-Net with several established approaches in domain-incremental learning, including Learning without Forgetting (LwF)~\cite{li2017learningforgetting}, Elastic Weight Consolidation (EWC)~\cite{Kirkpatrick_2017}, Experience Replay (ER)~\cite{chaudhry2019tinyepisodicmemoriescontinual}, Experience Replay with Asymmetric Cross-Entropy (ER-ACE)~\cite{caccia2022newinsightsreducingabrupt}, and Dark Experience Replay++ (DER++)~\cite{buzzega2020darkerexperiencegeneralcontinual} in Table~\ref{tab:overallresults}. We used the same CNN backbone across all methods for fair comparison. We observed that CLAD-Net outperforms all replay-free baselines in terms of final accuracy and forgetting across all three datasets. The results demonstrate that CLAD-Net achieves consistently strong performance across datasets with varying numbers of subjects (8 for PAMAP2 and DnSA, 15 for RealWorld). Replay-based methods achieve higher final accuracy (particularly on DnSA and RealWorld datasets), demonstrating the effectiveness of exemplar storage. However, CLAD-Net provides a privacy-preserving alternative that achieves competitive results without storing any user data. This distinction is particularly significant in real-world HAR applications where sensor data often contains sensitive personal information, including movement patterns and daily routines, that users may not consent to store indefinitely. Furthermore, CLAD-Net's exemplar-free approach eliminates the need for replay memory management and reduces storage requirements, making it more suitable for privacy-sensitive deployment scenarios such as healthcare monitoring and personal activity tracking. Importantly, as we show in Section \ref{sec:semisupervised}, CLAD-Net's robustness becomes particularly evident in label-scarce scenarios, where it outperforms replay-based methods by maintaining lower forgetting rates when labeled data is limited. Overall, CLAD-Net shows a superior trade-off between remembering old subjects and learning from new ones, without the need to include a separate memory in the architecture.

\begin{table*}[ht]
\centering
\caption{Comparison of CLAD-Net with baseline domain-incremental learning methods on PAMAP2, DnSA, and RealWorld datasets.}
\label{tab:overallresults}
\resizebox{\textwidth}{!}{%
\begin{tabular}{l|ccc|ccc|ccc}
\toprule
& \multicolumn{3}{c|}{\textbf{PAMAP2}} & \multicolumn{3}{c|}{\textbf{DnSA}} & \multicolumn{3}{c}{\textbf{RealWorld}} \\
\textbf{Model} & \textbf{FA} $\uparrow$ & \textbf{FM} $\downarrow$ & \textbf{LA} $\uparrow$ 
              & \textbf{FA} $\uparrow$ & \textbf{FM} $\downarrow$ & \textbf{LA} $\uparrow$ 
              & \textbf{FA} $\uparrow$ & \textbf{FM} $\downarrow$ & \textbf{LA} $\uparrow$ \\
\midrule
LwF                                        & 77.02          & 11.41         & 87.18          & 67.45          & 23.93          & 86.07         & 61.57          & 27.07         & 86.15 \\
EWC                                       & 69.63          & 28.87         & 98.50          & 74.04      & 25.34        & 99.38  & 53.76          & 45.18         & 98.94 \\
ER                                        & 83.47          & 14.45         & 97.92          & 89.62     & 9.62            & 99.23     & 75.23          & 22.87         & 98.10 \\
ER-ACE                                    & 83.85          & 14.20         & 98.05          & 89.95     & 9.35            & 99.40     & 75.65          & 22.45         & 98.25 \\
DER++                                     & 84.15          & 13.95         & {98.15}          & 90.25     & 9.10            & {99.55}     & 76.05          & 22.10         & {98.35} \\
\midrule
CLAD-Net & 80.78 & 11.05 & 91.22 & 81.14 & 8.68 & 89.55 & 64.34 & 24.34 & 87.51 \\
\bottomrule
\end{tabular}}
\end{table*}

To further evaluate CLAD-Net, we also compared it with ConvBoost~\cite{convboost}, a leading ensemble-based method for activity recognition that leverages diversity techniques like mixup augmentation, channel dropout, and dynamic data sampling. We experimented with three ConvBoost variants, each using a different backbone: a standard CNN, a ConvLSTM, and an attention-based RCNN model. As shown in Table~\ref{tab:conboostcomparison}, all three versions achieved high learning accuracy (91-94\%), often exceeding CLAD-Net's LA, suggesting that ConvBoost can effectively learn representations when trained on individual subjects. However, since none of these variants include explicit continual learning strategies, they suffered from notably higher forgetting rates compared to CLAD-Net across all three datasets. This comparison underscores that even high-performing HAR models may struggle in continual learning scenarios without mechanisms specifically designed to retain knowledge over time.

\begin{table*}[ht]
\centering
\caption{Comparison of CLAD-Net with ConvBoost variants on PAMAP2, DnSA, and RealWorld datasets.}
\label{tab:conboostcomparison}
\resizebox{\textwidth}{!}{%
\begin{tabular}{l|ccc|ccc|ccc}
\toprule
& \multicolumn{3}{c|}{\textbf{PAMAP2}} & \multicolumn{3}{c|}{\textbf{DnSA}} & \multicolumn{3}{c}{\textbf{RealWorld}} \\
\textbf{Model} & \textbf{FA} $\uparrow$ & \textbf{FM} $\downarrow$ & \textbf{LA} $\uparrow$ 
              & \textbf{FA} $\uparrow$ & \textbf{FM} $\downarrow$ & \textbf{LA} $\uparrow$ 
              & \textbf{FA} $\uparrow$ & \textbf{FM} $\downarrow$ & \textbf{LA} $\uparrow$ \\
\midrule
ConvBoost (CNN)        & 58.02 & 33.97 & 93.74 & 64.04 & 22.64 & 91.39 & 44.17 & 44.90 & 91.22 \\
ConvBoost (ConvLSTM)   & 54.89 & 36.19 & 93.34 & 60.58 & 25.17 & 91.28 & 40.93 & 47.17 & 90.86 \\
ConvBoost (Att. Model) & 69.59 & 22.74 & {94.10} & 65.26 & 21.23 & {91.40} & 55.25 & 33.75 & {91.51} \\
\midrule
CLAD-Net & 80.78 & 11.05 & 91.22 & 81.14 & 8.68 & 89.55 & 64.34 & 24.34 & 87.51 \\
\bottomrule
\end{tabular}}
\end{table*}

\subsection{Ablation Study of CLAD-Net's Components}

\subsubsection{Importance of each component}
To better understand the contribution of each component in CLAD-Net, we performed an ablation study by systematically removing or modifying key parts of the architecture. Table~\ref{tab:ablations} presents a summary of these ablations. The results demonstrate that both the self-supervised transformer and the knowledge distillation mechanism are essential for achieving high final accuracy and low forgetting across all three datasets. Specifically, removing the transformer leads to a substantial increase in forgetting, while omitting knowledge distillation also degrades overall performance.

\begin{table*}[ht]
\centering
\caption{Ablation study of CLAD-Net's components on PAMAP2, DnSA, and RealWorld datasets.}
\label{tab:ablations}
\resizebox{\textwidth}{!}{%
\begin{tabular}{l|ccc|ccc|ccc}
\toprule
& \multicolumn{3}{c|}{\textbf{PAMAP2}} & \multicolumn{3}{c|}{\textbf{DnSA}} & \multicolumn{3}{c}{\textbf{RealWorld}} \\
\textbf{Model} & \textbf{FA} $\uparrow$ & \textbf{FM} $\downarrow$ & \textbf{LA} $\uparrow$ 
              & \textbf{FA} $\uparrow$ & \textbf{FM} $\downarrow$ & \textbf{LA} $\uparrow$ 
              & \textbf{FA} $\uparrow$ & \textbf{FM} $\downarrow$ & \textbf{LA} $\uparrow$ \\
\midrule
w/o Transformer               & 77.02 & 11.41 & 87.18 & 67.45 & 23.93 & 86.07 & 61.57 & 27.07 & 86.15 \\
w/o Distillation              & 76.21 & 22.54 & 97.75 & 63.02 & 36.24 & 99.27 & 59.93 & 39.22 & 96.15 \\
\midrule
CLAD-Net & 80.78 & 11.05 & 91.22 & 81.14 & 8.68 & 89.55 & 64.34 & 24.34 & 87.51 \\
\bottomrule
\end{tabular}}
\end{table*}

\subsubsection{SSL Loss Functions}
We begin by evaluating the effect of different loss functions used to train the transformer. Specifically, we compare three widely adopted self-supervised learning losses: Barlow Twins~\cite{zbontar2021barlow}, SimCLR~\cite{chen2020simple}, and BYOL~\cite{grill2020bootstrap}. Barlow Twins encourages the embeddings from two views to be highly correlated across matching dimensions while minimizing redundancy across different dimensions by pushing the cross-correlation matrix toward the identity. SimCLR, on the other hand, uses contrastive learning to pull together representations of positive pairs (i.e., two augmented versions of the same input) while pushing apart those of different inputs. BYOL follows a teacher-student framework, where a momentum-updated target network guides a student network to produce consistent representations from two augmented views of the same input—without relying on negative samples.

\begin{table*}[ht]
\centering
\caption{Comparison of Self-Supervised Learning Loss Functions used within the transformer model on PAMAP2, DnSA, and RealWorld datasets.}
\label{tab:loss_functions}
\resizebox{\textwidth}{!}{%
\begin{tabular}{l|ccc|ccc|ccc}
\toprule
& \multicolumn{3}{c|}{\textbf{PAMAP2}} & \multicolumn{3}{c|}{\textbf{DnSA}} & \multicolumn{3}{c}{\textbf{RealWorld}} \\
\textbf{SSL Method} & \textbf{FA} $\uparrow$ & \textbf{FM} $\downarrow$ & \textbf{LA} $\uparrow$
                    & \textbf{FA} $\uparrow$ & \textbf{FM} $\downarrow$ & \textbf{LA} $\uparrow$ 
                    & \textbf{FA} $\uparrow$ & \textbf{FM} $\downarrow$ & \textbf{LA} $\uparrow$ \\
\midrule

SimCLR  & 76.85 & 14.43 & 90.86 & 78.21 & 13.57 & 91.79 &   60.87 &   29.68 &   88.82  \\
{Barlow Twins}  & 78.68 & 14.23 & 92.66 & 69.34 & 21.23 & 90.02 &   60.91 &   31.62 &   91.11 \\

BYOL        & 80.78 & 11.05 & 91.22 & 81.14 & 8.68 & 89.55 & 64.34 & 24.34 & 87.51 \\

\bottomrule
\end{tabular}}
\end{table*}

According to Table~\ref{tab:loss_functions} and as previously demonstrated in self-supervised continual learning works such as~\cite{fini2022selfsupervisedmodelscontinuallearners} and~\cite{pham2021dualnetcontinuallearningfast}, we also observed that the BYOL loss achieves superior performance in terms of final accuracy and forgetting measure on all three datasets. Moreover, this loss function is particularly well-suited for real-world applications due to its ability to avoid reliance on negative pairs or large batch sizes.

\begin{figure*}
    \centering\includegraphics[width=\linewidth]{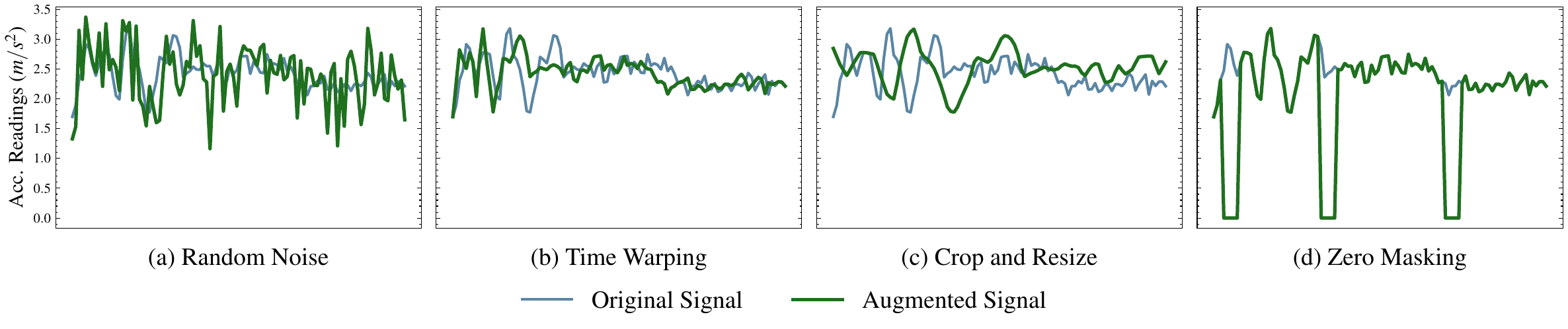}
    \caption{Visualization of four different time-series augmentation methods applied to a sample accelerometer signal.}
    \label{fig:augmethods}
\end{figure*}

\subsubsection{Augmentation Methods}
Before training the self-supervised model with the BYOL loss, two distinct views of the input must be generated using augmentation techniques designed for time-series data. We evaluated four such augmentation methods and selected the one that performed best for our model. Fig.~\ref{fig:augmethods} illustrates the effect of these four methods on a randomly chosen accelerometer time series. The corresponding performance results are summarized in Table~\ref{tab:augmentations}. As shown, the zero-masking method outperforms the others, particularly in terms of the forgetting ratio across all three datasets. Based on these results, we used this augmentation method to train CLAD-Net.

\begin{table*}[ht]
\centering
\caption{Comparison of Different Augmentation methods used within the transformer model on PAMAP2, DnSA, and RealWorld datasets.}
\label{tab:augmentations}
\resizebox{\textwidth}{!}{%
\begin{tabular}{l|ccc|ccc|ccc}
\toprule
& \multicolumn{3}{c|}{\textbf{PAMAP2}} & \multicolumn{3}{c|}{\textbf{DnSA}} & \multicolumn{3}{c}{\textbf{RealWorld}} \\
\textbf{Augmentation Method} & \textbf{FA} $\uparrow$ & \textbf{FM} $\downarrow$ & \textbf{LA} $\uparrow$
                    & \textbf{FA} $\uparrow$ & \textbf{FM} $\downarrow$ & \textbf{LA} $\uparrow$ 
                    & \textbf{FA} $\uparrow$ & \textbf{FM} $\downarrow$ & \textbf{LA} $\uparrow$ \\
\midrule
Random Noise         & 77.66  & 13.30 & 90.09 & 78.53 & 12.79 & 91.25 & 60.42 & 28.95 & 87.67 \\
Time Warping  & 77.12  & 14.13 & 90.62 & 77.43 & 13.70 & 91.07 & 60.64 & 29.16 & 88.15 \\
{Crop and Resize}  & {77.38} & {17.13}  &  {91.06} &  69.15 & 21.13 & 91.38 & 60.91 & 31.62 & 91.11
 \\
Zero Masking  & 80.78 & 11.05 & 91.22 & 81.14 & 8.68 & 89.55 & 64.34 & 24.34 & 87.51 \\

\bottomrule
\end{tabular}}
\end{table*}

\subsubsection{Cross-Attention vs. Self-Attention}
Finally, to demonstrate the effectiveness of cross-attention, we compared our transformer model using cross-attention blocks with an identical model that relied solely on self-attention, where each body part's input attended only to itself. Both models were trained under the same self-supervised and continual learning setup to ensure a fair comparison. As shown in Table~\ref{tab:attentions}, the cross-attention model outperformed its self-attention counterpart on final accuracy (FA) and forgetting measure (FM), demonstrating that modeling inter-sensor relationships yields richer representations and better knowledge preservation across subjects.

\begin{table*}[ht]
\centering
\caption{Comparison of various attention mechanisms used within the transformer model on PAMAP2, DnSA, and RealWorld datasets.}
\label{tab:attentions}
\resizebox{\textwidth}{!}{%
\begin{tabular}{l|ccc|ccc|ccc}
\toprule
& \multicolumn{3}{c|}{\textbf{PAMAP2}} & \multicolumn{3}{c|}{\textbf{DnSA}} & \multicolumn{3}{c}{\textbf{RealWorld}} \\
\textbf{Attention Mechanism} & \textbf{FA} $\uparrow$ & \textbf{FM} $\downarrow$ & \textbf{LA} $\uparrow$
                    & \textbf{FA} $\uparrow$ & \textbf{FM} $\downarrow$ & \textbf{LA} $\uparrow$ 
                    & \textbf{FA} $\uparrow$ & \textbf{FM} $\downarrow$ & \textbf{LA} $\uparrow$ \\
\midrule
Self Attention         & 77.59  & 14.47 & 91.54 & 75.90 & 15.06 & 90.78 & 62.38 & 29.71 & 90.58 \\
Cross Attention  & 80.78 & 11.05 & 91.22 & 81.14 & 8.68 & 89.55 & 64.34 & 24.34 & 87.51 \\
\bottomrule
\end{tabular}}
\end{table*}

\subsection{Semi-Supervised Setting}
\label{sec:semisupervised}

In practical scenarios, datasets are rarely fully labeled. Instead, models usually have to deal with a mix of labeled and unlabeled data. This setup often leads to lower overall accuracy, but how much a model forgets in such cases can vary quite a bit.

To explore this, we ran one last set of experiments where we randomly removed labels from most of the data, keeping only $\varphi = 10\%$ and $\varphi = 20\%$ of it labeled. We tested several models under these conditions. In our method, the transformer still uses all the data because its self-supervised training doesn't rely on labels. But the classifier part only sees the labeled portion during training.

\begin{figure*}
    \centering
    \includegraphics[width=\linewidth]{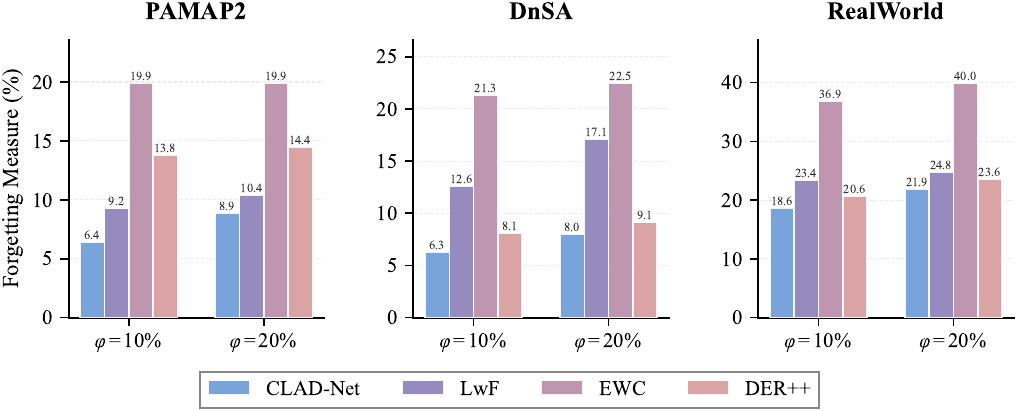}
    \caption{Forgetting comparison across different models under 10\% and 20\% label availability on the PAMAP2, DnSA and RealWorld datasets. CLAD-Net consistently shows lower forgetting than other methods.}
    \label{fig:labels}
\end{figure*}

Fig.~\ref{fig:labels} shows the forgetting measure for CLAD-Net, LwF, EWC, and DER++ under both label settings. As the figure shows, our model consistently forgets less than the others. We hypothesize that this is mostly due to the self-supervised part of CLAD-Net, especially since the LwF version (basically our model without the SSL module) performs worse in every case.\looseness=-1

%% file: Sections/05_conclusion.tex
\section{Conclusion and Future Work}
In this work, we address the challenge of domain-incremental learning in HAR for healthcare monitoring applications. Reliable activity recognition across diverse patient populations is essential for clinical decision support systems, rehabilitation tracking, and long-term patient monitoring. We highlight that transferring a model trained on one patient to data from a new patient often leads to catastrophic forgetting due to individual variations in movement patterns, sensor placement, and activity execution styles. This degradation stems from distributional shifts in sensor readings, as different individuals tend to perform the same activity in distinct ways, as illustrated in Fig.~\ref{fig:forgetting_intro}.

\begin{figure}
    \centering
    \includegraphics[width=\linewidth]{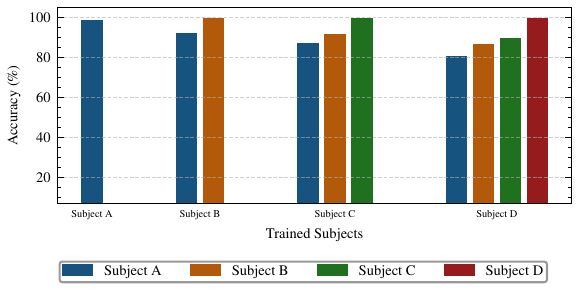}
    \caption{CLAD-Net performance on the first four subjects from the PAMAP2 dataset.}
    \label{fig:forgetting-conclusion}
\end{figure}

To mitigate this issue, we proposed CLAD-Net, a dual-component framework designed for continual activity recognition in multi-sensor wearable settings. CLAD-Net combines a self-supervised transformer with cross-attention and a supervised CNN classifier, linked through knowledge distillation, to address the problem of subject-incremental learning. In Fig.~\ref{fig:forgetting-conclusion}, we apply CLAD-Net to the same sequence of subjects as in Fig.~\ref{fig:forgetting_intro}, showing that our model substantially reduces forgetting compared to naive sequential training and distillation-free approaches. Our experimental results confirm that both the self-supervised representation learner and the distillation mechanism are critical for maintaining performance across sequential tasks.

We conducted extensive evaluations on three benchmark datasets, PAMAP2~\cite{pamap2_physical_activity_monitoring_231}, DnSA~\cite{daily_and_sports_activities_256}, and RealWorld~\cite{sztyler2016realworld}, where CLAD-Net outperformed replay-free baselines (LwF and EWC) and achieved competitive forgetting measures while offering a privacy-preserving alternative to replay-based methods (ER, ER-ACE, DER++). Notably, in label-scarce scenarios, CLAD-Net demonstrated superior forgetting resistance compared to all baselines, highlighting the value of self-supervised learning when labeled data is limited. Ablation studies further demonstrated that removing either the self-supervised transformer or the knowledge distillation module leads to a noticeable drop in performance. In addition, we compared cross-attention with self-attention and showed that modeling interactions across different body parts leads to better final accuracy and lower forgetting compared to having each body part attend only to itself.

For future work, we plan to extend CLAD-Net to support fully class-incremental learning settings and incorporate additional sensor modalities for more dynamic, online learning scenarios. Another important direction is to evaluate the scalability and resilience of these methods when exposed to even larger numbers of tasks. In particular, we are interested in understanding how the model behaves as the number of subjects increases beyond the 15 in our largest dataset. Overall, our results emphasize the promise of integrating self-supervised learning with cross-attention, knowledge distillation, and multi-sensor fusion to build more scalable and robust HAR systems for healthcare applications.